\documentclass{article}

\pdfoutput=1
\usepackage{ijcai17}
\usepackage{times}

\usepackage{amsmath,amssymb,amstext,amsthm} 
\usepackage[pdftex]{graphicx} 
\usepackage{mathtools}
\usepackage{subcaption}
\usepackage{caption}
\usepackage{algorithm}
\usepackage{algorithmic}
\usepackage{multirow}
\usepackage{ctable}
\usepackage{hyperref}

\DeclarePairedDelimiter\abs{\lvert}{\rvert}%
\DeclarePairedDelimiter\norm{\lVert}{\rVert}%
\newtheorem{theorem}{Theorem}[section]
\newtheorem{corollary}{Corollary}[theorem]
\newtheorem{lemma}[theorem]{Lemma}

\DeclareMathOperator*{\argmin}{arg\,min}
\DeclareMathOperator*{\argmax}{arg\,max}

\newif\iffull
\fulltrue 

\newif\ifshowauthor
\showauthortrue

\title{Projection Free Rank-Drop Steps}
\ifshowauthor
\author{Edward Cheung \phantom{{}\and{}} Yuying Li\\ 
Cheriton School of Computer Science, University of Waterloo, Waterloo, Canada  \\
\{eycheung, yuying\}@uwaterloo.ca
}
\fi
\begin{document}

\maketitle

\begin{abstract}
The Frank-Wolfe (FW) algorithm has been widely used in solving nuclear norm constrained problems, since it does not require projections. However, FW often yields high rank intermediate iterates, which can be very expensive in time and space costs for large problems. To address this issue, we propose a rank-drop method for nuclear norm constrained problems. The goal is to generate descent steps that lead to rank decreases, maintaining low-rank solutions throughout the algorithm. Moreover, the optimization problems are constrained to ensure that the rank-drop step is also feasible and  can be readily incorporated into a projection-free minimization method, e.g., FW. We demonstrate that by incorporating rank-drop steps into the FW algorithm, the rank of the solution is greatly reduced compared to the original FW or its common variants.
\end{abstract}

\section{Introduction}

The Frank-Wolfe algorithm has been widely used for many machine learning applications due to its projection-free property, particularly when linear minimization on the domain is easy but projection is difficult. A particularly interesting problem in machine learning is the nuclear norm constrained problem,
\begin{equation}
\label{prob:nn_reg}
\min_{X \in \mathbb{R}^{m\times n}} f(X) ~ s.t. ~ \norm{X}_{NN} \leq \delta
\end{equation}
where $\norm{\cdot}_{NN}$ is the nuclear norm, which is a typical convex relaxation for rank constrained optimization problems \cite{fazel2001rank}. Common applications of nuclear-norm constrained problems are matrix completions, multivariate regression, multi-task learning, and clustering with missing information. For \eqref{prob:nn_reg}, a projection operation onto the nuclear norm ball will require a full singular value decomposition (SVD), which can be too expensive to perform at each iteration, but is required by methods such as the projected gradient descent.

Without any projection, the FW iterate is guaranteed to be feasible.  For \eqref{prob:nn_reg}, the linear subproblem used by FW only requires computing the  singular vector pair corresponding to the largest singular value of the gradient at each iteration \cite{jaggi2013revisiting}. This is significantly cheaper than computing the full SVD when the dimension of the matrix is large. A challenge  when using FW for \eqref{prob:nn_reg} is that the nuclear norm ball is a convex hull of an infinite number of rank-one matrices, referred to as the \emph{atoms} of the set of feasible points. Since an atom is added at each iteration, the solution is expressed as a convex combination of an arbitrarily large atomic set, which lacks  the crucial low-rank property we desire from nuclear norm constrained problems.

While other methods exist for solving instances of \eqref{prob:nn_reg}, e.g., Active Subspace Selection \cite{hsieh2014nuclear} or Redistributing Nonconvex Regularizers \cite{yao2016efficient}, implementations for these methods  utilize specific properties of $f(\cdot)$. In \cite{yao2016efficient}, it appears that for each choice of $f(\cdot)$, a specialized solver must be written, and in \cite{hsieh2014nuclear}, knowledge of special structures using $\nabla f(\cdot)$ and $\nabla^{2} f(\cdot)$ is necessary to use the approach efficiently. In contrast, the FW can be readily used for a general $f(\cdot)$ without special modifications.

One of our main goals is to devise a new way to efficiently generate descent feasible steps which also decrease the rank of the current iterate when solving \eqref{prob:nn_reg} without assuming a specific form for $f(\cdot)$. 
We propose a new nonconvex optimization formulation to determine the steepest descent rank-drop steps. By further considering the interior rank-drop step and exterior rank-drop step cases separately, we obtain formulations which lead to efficient feasible rank-drop step computation.

We establish theoretical properties of the proposed formulations. In addition we demonstrate computationally that the proposed rank-drop FW method can obtain much lower rank solutions than previous FW alternatives, greatly improving computational efficiency.

\subsection{Notation}
In this paper,  $\sigma_{i}(A)$ denotes the $i^{\textsuperscript{th}}$ singular value of $A$, with $\sigma_{1} \geq ... \geq \sigma_{\min\{m,n\}}$. For a matrix $A = [a_{ij}] \in \mathbb{R}^{m \times n}$,  $\norm{A}_{NN} \coloneqq \sum_{i}\sigma_{i}(A)$ denotes the nuclear norm and $\norm{A}_{sp} \coloneqq \sigma_{1}(A)$ denotes the spectral norm. Let $\mathcal{B}_{\norm{\cdot}}(A, \epsilon) \coloneqq \{X \in \mathbb{R}^{m\times n}: \norm{A - X} \leq \epsilon\}$ be the \emph{closed} norm-ball with a specified norm. We use the term \emph{thin SVD} to mean the SVD with strictly positive singular values, i.e., if $A = U\Sigma V^{\top}$ is a thin SVD and $rank(A) = r$, then $U \in \mathbb{R}^{m \times r}, \Sigma \in \mathbb{R}^{r \times r}$, and $V \in \mathbb{R}^{n \times r}$. Finally, $\langle A, B \rangle \coloneqq \text{tr}(A^{\top}B)$ denotes the usual trace inner product.

\section{The Frank-Wolfe Algorithm and Away Step}
\label{background}
We briefly review FW along with its common variants. Here FW is described for a general problem,
\begin{equation}
    \min_{x \in \mathcal{S}}~f(x)
\end{equation}
where $\mathcal{S}$ is a closed and bounded convex set.

The FW algorithm can be summarized in Algorithm \ref{vanilla-fw}. We denote the FW direction as $d_{fw} \coloneqq s_k - x_{k}$. 
\begin{algorithm}  
    \caption{Frank-Wolfe (\textbf{FW})}
  \label{vanilla-fw}
  \begin{algorithmic}
  \STATE Let $x_{0} \in \mathcal{S}$
  \FOR{$k = 0...$}
  \STATE $s_{k} \gets \argmin_{s \in \mathcal{S}} \langle s, \nabla f(x_{k}) \rangle$
  \STATE $d_{fw} \gets s_{k} - x_{k}$
  \STATE $x_{k+1} \gets x_{k} + \tau d_{fw}$ for $\tau \in [0, 1]$
  \ENDFOR 
\end{algorithmic}
\end{algorithm}

There are many choices for the stepsize $\tau$ which  ensure convergence. We will assume in this paper that the optimal step-size is used since the stepsize choice is not the focus of this work. 

Away steps has been introduced in \cite{wolfe1970convergence} and analyzed in \cite{guelat1986some} to move away from ``bad atoms'' to accelerate convergence. The away step method maintains a set of active atoms, $\mathcal{A}_{k}$, such that the current iterate $x_k$ is decomposed into an \emph{atomic decomposition}, i.e., $x_{k} = \sum_{a_{i} \in \mathcal{A}_{k}}\alpha_{a_{i}}a_{i}$, where the weight satisfy $\alpha_{a_{i}} > 0$ and $\sum_{a_{i} \in \mathcal{A}_{k}} \alpha_{a_{i}} \leq 1$.

The away step provides a direction that moves away from one of the active atoms which yields a better descent direction than the current FW direction. This direction solves the following optimization problem,
\begin{equation}
\label{eqn:away_step}
v_{k} \coloneqq \argmin_{v \in \mathcal{A}_{k}}  \langle \nabla f(x_{k}), x_{k} -  v_{k}\rangle
\end{equation}
To ensure convergence, the away direction $d_{away} \coloneqq x_{k} - v_{k}$ is only taken 
if $\langle \nabla f(x_{k}), d_{away} \rangle \leq \langle \nabla f(x_{k}), d_{fw} \rangle$. 
Algorithm \ref{away_step} summarizes the away step FW method  (AFW) in \cite{lacoste2015global}.

\begin{algorithm}
  \caption{(Atomic) Away Steps Frank-Wolfe (\textbf{AFW})}  
  \label{away_step}
  \begin{algorithmic}
  \STATE Let $x_{0} \in \mathcal{S}$ and $\mathcal{A}_{0} \gets \{x_{0}\}$.
  \STATE Initialize $\alpha_{s} \gets 0, \forall s \in \mathcal{S}$.
  \FOR{$k = 0...$}
  \STATE $s_{k} \gets \argmin_{s \in \mathcal{S}} \langle s, \nabla f(x_{k}) \rangle, d_{fw} \gets s_{k} - x_{k}$
  \STATE $v_{k} \gets \argmax_{v \in \mathcal{A}_{k}} \langle v, \nabla f(x_{k}), d_{away} \gets x_{k} - v_{k}$
  \IF{$\langle -\nabla f(x_{k}), d_{fw} \rangle \geq \langle -\nabla f(x_{k}), d_{away} \rangle$}
    \STATE $d_{k} \gets d_{fw}$    
    \STATE $\tau^{*} \gets \argmin_{\tau \in [0, 1]} f(x_{k} + \tau d_{k})$
    \STATE $\alpha_{a_{i}} \gets (1-\tau^{*})\alpha_{a_{i}}, \forall a_{i} \in \mathcal{A}_{k}$
    \STATE $\alpha_{s_{k}} \gets \alpha_{s_{k}} + \tau^{*}, \mathcal{A}_{k} \gets \mathcal{A}_{k} \cup \{s_{k}\}$
  \ELSE
    \STATE $d_{k} \gets d_{away}$
    \STATE $\tau^{*} \gets \argmin_{\tau \in \left[0, \frac{\alpha_{v_{k}}}{1 - \alpha_{v_{k}}}\right]} f(x_{k} + \tau d_{k})$
    \STATE $\alpha_{v_{i}} \gets (1+\tau^{*})\alpha_{v_{i}}, \forall v_{i} \in \mathcal{A}_{k}$
    \STATE $\alpha_{v_{k}} \gets \alpha_{v_{k}} - \tau^{*}$
  \ENDIF  
  \STATE $x_{k+1} \gets x_{k} + \tau^{*}d_{k}$  
  \STATE $\mathcal{A}_{k+1} \gets \{ a \in \mathcal{A}_{k} : \alpha_{a} > 0\} $
  \ENDFOR
\end{algorithmic}
\end{algorithm}

\subsection{In-Face Steps}
In \cite{freund2015extended},  \emph{In-Face Steps} are proposed to generalize  away-steps. Instead of generating descent directions by moving away from bad atoms, the in-face step selects the best descent direction along the minimal face containing the current iterate. By aiming to move along the minimal face  to the boundary of the feasible region to reach a lower dimensional face, the algorithm explicitly maintains a low-rank structure without compromising convergence. For \eqref{prob:nn_reg}, as in \cite{freund2015extended}, the minimal face $\mathcal{F}(X_{k})$ of $\mathcal{B}_{NN}(0, \delta)$ containing a point $X_{k}$  is given by the set,
\begin{equation}
\label{def:minimal_face}
\mathcal{F}(X_{k}) = \begin{cases}
\mathcal{B}_{NN}(0, \delta) \text{,~~ when $\norm{X_{k}} < \delta$}\\
UMV^{\top}, \text{    ~~~~~otherwise, }
\end{cases}
\end{equation}
where $X_{k}$ has a thin $r$ rank  SVD, $U\Sigma V^{\top}$,  $M$ is a real positive semidefinite matrix with $\text{tr}(M) = \delta$.

In particular, for comparison, we consider the In-Face step  from the \emph{Away-step Strategy} described in \cite{freund2015extended}, which uses the following direction,
\begin{equation}
\label{eqn:in-face}
Z_{k} \coloneqq \argmax_{Z \in \mathcal{F}(X_{k})} \langle \nabla f(X_{k}), Z \rangle.
\end{equation}

\section{Rank-Drop Steps}
\label{rd_steps}
\begin{figure}
\centering
\includegraphics[width=0.45\textwidth]{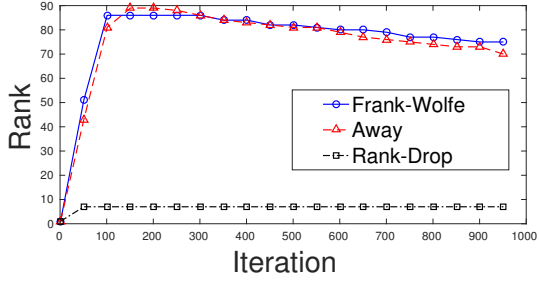}
\caption{\small The Frank-Wolfe and Away-Step methods reach very high rank solutions at convergence for the MovieLens 100k dataset with $\delta = 1$. The rank of the iterate becomes very large before the rank begins to decrease very slowly. In contrast, the proposed rank-drop FW method maintains very low-rank intermediate solutions throughout the entire computation. The objective values at termination are 1905.4, 1905.4, 1905.6 for FW, Away Step, Rank-Drop respectively.}
\label{fig:rank_drop_demo}
\end{figure}

One significant issue with the FW method for nuclear norm constrained problems is that the rank of the intermediate solution often steadily increases \cite{freund2015extended}, and frequently yielding a high rank solution upon termination. In Figure \ref{fig:rank_drop_demo}, we highlight this phenomenon. For very large problems, we cannot store the matrix $X_{k}$, and instead, only maintain the rank-one factors. Thus, computing the gradient at each iteration can be computationally prohibitive.  For matrix completion, as an example, if $\Omega$ is the set of observed entries, then forming the gradient requires at least $\abs{\Omega} \cdot r$ inner product calculations. Since the number of observed entries, $\abs{\Omega}$, is typically very large, any increase in the rank of the solution greatly increases the computational time in each iteration.

To address this issue, we propose to search over the set of rank-one matrices which decrease the rank of the iterate by one. The proposed optimization formulation for determining such a rank-one matrix is motivated by the following theorem.

\begin{theorem}[{\cite{egervary1960rank}}]
\label{thm:rank_drop}
Let $u \in \mathbb{R}^{m}$,  $v \in \mathbb{R}^{n}$,  $A \in \mathbb{R}^{m\times n}$, and  $B = A - \sigma^{-1}uv^{\top}$.  Then $rank(B) = rank(A) - 1$  if and only if there are vectors $x \in \mathbb{R}^{n}$ and $y \in \mathbb{R}^{m}$ such that $u = Ax$, $v = A^{\top}y$ and $\sigma = y^{\top}Ax \neq 0$.
\end{theorem}

\subsection{The Rank-Drop Optimization Framework}
Assume that $rank(X_{k}) > 1$. Our goal is to determine a rank-drop step  which reduces the rank of  $X_k$ but has a comparable objective value. We restrict our attention to the rank-one matrices that can decrease the rank of a matrix; let ${\mathcal R}(A)$ denote, for a matrix $A$, the set of \emph{rank-drop steps}, which are these rank-one matrices, 
\begin{align}
\label{def:orig_rd}
\begin{split}
\mathcal{R}(A) \coloneqq& \{\sigma^{-1}uv^{\top}: \exists x \in \mathbb{R}^{n}, y \in \mathbb{R}^{m} \\
& s.t. ~u = Ax, v = A^{\top}y, \sigma = y^{\top}Ax \neq 0\}.
\end{split}
\end{align}
Since $u$ and $v$ must be in the column and row spans of $A$ respectively, Lemma \ref{lem:rewrite_rd} shows that the set of rank-drop steps can be expressed in a more concise form.
\begin{lemma}\footnote{Due to space limitations, proofs will be omitted. The proofs will be added in an appendix at \url{https://arxiv.org/abs/1704.04285}.}
\label{lem:rewrite_rd}
Let $A \in \mathbb{R}^{m \times n}$ with a thin SVD $A = U\Sigma V^{\top}$. Then we can rewrite the set of rank-drop steps as, 
\begin{equation}
\mathcal{R}(A) = \left\{ \frac{Ust^{\top}V^{\top}}{s^{\top}\Sigma^{-1}t}, s^{\top}\Sigma^{-1}t > 0 \right\}.
\end{equation}
\end{lemma}

Let $Z_{k} \in \mathcal{R}(X_{k})$ and $\hat{Z}_{k} = Z_{k}/\norm{Z_{k}}_{NN}$. Then, similar to  away steps, we consider iterates in the following form,
\begin{equation}
\label{def:rd_iterates}
X_{k+1} \coloneqq X_{k} + \tau D_{rd}, \quad \text{ where } D_{rd} \coloneqq X_{k} - \delta\hat{Z}_{k}.
\end{equation}

Our goal is to find a rank-drop step that results in a feasible iterate which best minimizes the objective. Since not all rank-drop steps lead to feasible iterates and testing for feasibility at each iteration is computationally expensive,  we want to establish a verifiable sufficient condition for rank-drop steps that ensure feasibility. In Lemma \ref{lem:radius_bound},  we first establish a lower bound on the nuclear norm of the rank-drop step.

\begin{lemma}
\label{lem:radius_bound}
Let $A \in \mathbb{R}^{m \times n}$ with rank $r$ and let $Z \in \mathcal{R}(A)$ be an arbitrary rank-drop step. Then $\norm{Z}_{NN} \geq \sigma_{r}(A)$.
\end{lemma}

To ensure feasibility of the iterate after a rank-drop step, we consider two distinct cases. Let $\kappa(X_{k})$ be half of the distance between $X_k$ and the boundary of the nuclear norm ball,
\begin{equation}
\label{def:kappa}
\kappa(X_{k}) \coloneqq \frac{\delta - \norm{X_{k}}_{NN}}{2}
\end{equation}
We consider  $\kappa(X_{k}) \geq \sigma_{r}(X_{k})$,
the \emph{interior case}, and
$\kappa(X_{k}) < \sigma_{r}(X_{k})$, the \emph{exterior case}.  For each case, we will motivate a rank-drop step formulation to guarantee feasibility of the next iterate and provide a method for its computation.

\subsection{The Interior Rank-Drop Problem}
 Assume that $\kappa(X_{k}) \geq \sigma_{r}(X_{k})$, where $\kappa(X_{k})$ is defined in \eqref{def:kappa}. Theorem \ref{thm:maintain_cvx} below establishes an easily verifiable condition  guaranteeing that  $X_{k+1}$, after taking the rank-drop step, remains in the interior of the nuclear ball.

\begin{theorem}
\label{thm:maintain_cvx}
Let $X_{k} \in \mathbb{R}^{m \times n}$ with $\norm{X_{k}}_{NN} < \delta$, $Z_{k} \in \mathcal{R}(A)$, and $\hat{Z}_{k} = Z_{k}/\norm{Z_{k}}_{NN}$. If $\norm{Z_{k}} \leq (\delta - \norm{X_{k}}_{NN})/2$, then  $X_{k+1} = X_{k} + \tau(X_{k} - \delta\hat{Z}_{k}) \in \mathcal{B}_{NN}(0, \delta)$
for every $\tau \in [0,\tau^{*}]$, where $\tau^{*} = \norm{Z_{k}}_{NN}/(\delta - \norm{Z_{k}}_{NN})$. Moreover, if $\tau = \tau^{*}$, then $rank(X_{k+1}) = rank(X_{k}) - 1$.
\end{theorem}

Theorem \ref{thm:maintain_cvx} motivates the following formulation,
\begin{align}
\begin{split}
\label{intCase}
\min_{Z} \quad & \langle - Z, \nabla f(X_{k}) \rangle\\
~ s.t. \quad &~ Z \in \mathcal{B}_{NN}(0, \kappa(X_{k})) \cap \mathcal{R}(X_{k}).
\end{split}
\end{align}

Problem  \eqref{intCase} determines, in the interior case, the best  rank-drop step for solving \eqref{prob:nn_reg} based on the first order approximation to the objective. The constraint on $Z$ guarantees that the next iterate is feasible if a rank-drop step is taken.

\begin{theorem}
\label{thm:feasible}
If $rank(X_{k}) \geq 1$ with $\sigma_{r}(X_{k}) \leq \kappa(X_{k})$, then the feasible region for \eqref{intCase} is non-empty. 
\end{theorem}

Assume that  a thin SVD for $X_k$,  $X_{k} = U\Sigma V^{\top}$, is given.  Using Lemma \ref{lem:rewrite_rd}, the constraint in \eqref{intCase} can be made explicit, 
\begin{align}
\begin{split}
\label{prob:fractional}
\min_{s,t \in \mathbb{R}^{r}} &\left\langle X_{k} - \frac{Ust^{\top}V^{\top}}{s^{\top}\Sigma^{-1}t}, \nabla f(X_{k}) \right\rangle \\
s.t.~ &\frac{Ust^{\top}V^{\top}}{s^{\top}\Sigma^{-1}t} \in \mathcal{B}_{NN}(0, \kappa(X_{k}))
\\& s^{\top}\Sigma^{-1}t > 0.
\end{split}
\end{align}

To make \eqref{prob:fractional} more amenable to computation, we remove the fraction using the normalization constraint  $s^{\top}\Sigma^{-1}t = \kappa(X_{k})^{-1}$
and formulate the problem equivalently as follows,
\begin{align}
\begin{split}
\label{prob:rank_drop}
    \min_{s, t \in \mathbb{R}^{r}}~&q(s,t) := \langle \nabla f(X_{k}), -\kappa(X_{k})Ust^{\top}V^{\top} \rangle\\
    s.t.~&s^{\top}\Sigma^{-1}t = \kappa(X_{k})^{-1}\\
    ~& \norm{s}_{2} = 1, ~\norm{t}_{2} \leq 1
    \end{split}
\end{align}

Note that the constraints in \eqref{prob:rank_drop} also ensure that $s$ and $t$ cannot be rescaled to obtain a different solution yielding an identical rank-drop step. 
The equivalence of \eqref{intCase} and \eqref{prob:rank_drop}  is formally established in
Theorem \ref{thm:equv}.

\begin{theorem}\label{thm:equv}
 If $X_{k}$ is a feasible point of \eqref{prob:nn_reg} and $\kappa(X_{k}) \geq \sigma_{r}(X_{k})$, then an optimal solution to \eqref{prob:rank_drop} is an optimal solution to \eqref{prob:fractional}. Moreover, an optimal to \eqref{prob:fractional} can always be rescaled into an optimal solution to \eqref{prob:rank_drop}.
 \end{theorem}

\subsection{The Exterior Rank-Drop Problem }
Now consider the case $\kappa(X_{k}) < \sigma_{r}(X_{k})$. If $\kappa(X_{k})$ is relatively large and $X_k$ is not nearly rank deficient, rank-drop is less important. If $\kappa(X_{k})$ is small, however, then $X_k$ is close to the boundary of the nuclear ball. In this case, we are interested in rank-drop descent directions that move inside the nuclear norm ball. We establish the following theorems to facilitate formulating appropriate optimization problems for this case. Subsequently $\partial g(X)$ denotes the subdifferential of $g(X)$ at $X$.  

\begin{theorem}
\label{thm:descent}
Let $X_{k} \in \mathcal{B}_{NN}(0, \delta)$ have the thin SVD $X_{k} = U\Sigma V^{\top}$. Define $D_{k} = X_{k} - \delta\hat{Z}_{k}$ with $Z_{k} \in \mathcal{R}(X_{k})$ and $\hat{Z}_{k} = Z_{k}/\norm{Z_{k}}_{NN} = Ust^{\top}V^{\top}$, for some $s$ and $t$ with $\norm{s} = \norm{t} = 1$. Then,
\begin{equation}
  \left(  \max_{G \in \partial\norm{X_{k}}_{NN}}\langle D_{k}, G \rangle\right) \leq 0,
\end{equation}
if and only if $ \delta s^{\top}t \geq \norm{X_{k}}_{NN}$.
\end{theorem}

Following Theorem \ref{thm:descent}, $D_{k}$ will be a descent direction for the nuclear norm at $X_{k}$ if $\delta s^{\top}t \geq \norm{X_{k}}_{NN}$.

When $X_{k}$ is on the boundary of the nuclear norm ball, Corollary \ref{cor:sym} below states  a simpler and useful characterization.

\begin{corollary}
\label{cor:sym}
Let $X_{k}$ have the thin SVD $X_{k} = U\Sigma V^{\top}$ with $\norm{X_{k}}_{NN} = \delta$. Define $D_{k} = X_{k} - \delta\hat{Z}_{k}$ with $Z_{k} \in \mathcal{R}(X_{k})$ and $\hat{Z}_{k} = Z_{k}/\norm{Z_{k}}_{NN} = Ust^{\top}V^{\top}$, for some $s$ and $t$ with $\norm{s}_{2} = \norm{t}_{2} = 1$. Then,
\begin{equation}
   \left( \max_{G \in \partial\norm{X_{k}}_{NN}}\langle D_{k}, G \rangle \right)  \leq 0 
\end{equation}
if and only if  $s = t$.
\end{corollary}

When $X_{k}$ is on the boundary of the nuclear ball, $\kappa(X_{k})=0$.
Lemma \ref{lm:bdnn} below shows 
that $X_{k}$ is close to the boundary when either $r$ is large or when $\sigma_{r}(X_{k})$ is small, assuming $\kappa(X_{k}) < \sigma_{r}(X_{k})$.

\begin{lemma}\label{lm:bdnn}
If $\kappa(X_{k}) < \sigma_{r}(X_{k})$, then $\norm{X_{k}}_{NN} > \frac{r}{r+2}\delta$.
\end{lemma}

For the exterior case, $\kappa(X_{k}) < \sigma_{r}(X_{k})$, we are mostly interested in the situation when $X_{k}$ is close to the boundary of the  nuclear norm ball. Consequently, we motivate the next formulation assuming $\norm{X_{k}}_{NN} = \delta$ as a good approximation.

When $X_{k}$ is on the boundary of the nuclear norm ball, we consider all directions which move into the nuclear norm ball, hence feasible directions, as candidate solutions.
From Corollary \ref{cor:sym}, this can be reduced to searching over directions $D_{k} = X_{k} - \delta Uss^{\top}V^{\top}$. Let $W = U^{\top}\nabla f(X_{k})V$. Note that $\frac{1}{2}s^{\top}(W^{\top} + W)s = \langle \nabla f(X_{k}), Uss^{\top}V^{\top} \rangle$. Hence, in the exterior case, we solve the following optimization problem,

\begin{align}
\begin{split}
    \label{ann_rank_drop}
    \max_{s \in \mathbb{R}^{r}}~ \frac{1}{2}\frac{s^{\top}(W^{\top} + W)s}{s^{\top}\Sigma^{-1}s} ~ s.t. ~  \norm{s}_{2} = 1.
    \end{split}
\end{align}

We note that \eqref{ann_rank_drop} is a generalized eigenvalue problem, which  can be further reduced to a standard eigenvalue problem since $\Sigma^{-1}$ is a nonsingular diagonal matrix.

Theorem \ref{thm:ext_feasible} below  establishes that, using \eqref{ann_rank_drop}, the required step-size to decrease the rank of the current iterate also  guarantees  that the next iterate is feasible. 

\begin{theorem}
\label{thm:ext_feasible}
Let $s$ be an optimal solution to $\eqref{ann_rank_drop}$ and let $D_{k} = X_{k} - \delta Uss^{\top}V^{\top}$, where $X_{k} = U\Sigma V^{\top}$ is a thin SVD with $\norm{X_{k}}_{NN} \leq \delta$. If $\tau^{*} = (\delta s^{\top}\Sigma^{-1}s - 1)^{-1}$ and $X_{k+1} = X_{k} + \tau^{*}D_{k}$, then $rank(X_{k+1}) = rank(X_{k}) - 1$ and $\norm{X_{k+1}}_{NN} \leq \delta$.
\end{theorem}
We observe that  $\tau^{*}$ matches the upper bound on the step-size for in-face steps when an away-step method is used and when $\norm{X_{k}} = \delta$. 

\subsubsection{Comparison with In-Face Steps}
Lemma \ref{lem:int_if} and \ref{lem:ext_if} below show that
the rank-drop steps derived from \eqref{prob:rank_drop} and \eqref{ann_rank_drop} lie on the minimal face of $\mathcal{B}_{NN}(0, \delta)$ containing $X_{k}$, denoted as $\mathcal{F}(X_{k})$. 
\begin{lemma}
\label{lem:int_if}
Suppose $\norm{X_{k}}_{NN} < \delta$ and $\kappa(X_{k}) \geq \sigma_{r}(X_{k})$, with thin SVD $X_{k} = U\Sigma V^{\top}$. Then $Z = \kappa(X_{k}) Ust^{\top}V^{\top} \in \mathcal{F}(X_{k})$, where $(s,t)$ is the solution to \eqref{prob:rank_drop}.
\end{lemma}

\begin{lemma}
\label{lem:ext_if}
Suppose $\kappa(X_{k}) < \sigma_{r}(X_{k})$, with thin SVD $X_{k} = U\Sigma V^{\top}$. Then $Z = \delta Uss^{\top}V^{\top} \in \mathcal{F}(X_{k})$, where $s$ is the solution to \eqref{ann_rank_drop}.
\end{lemma}
We emphasize, however, that the rank-drop steps are constructed
explicitly to decrease the rank of the current iterate. Specifically, we highlight the following differences between rank-drop steps and the in-face direction \eqref{eqn:in-face} suggested in \cite{freund2015extended}. Firstly, when the iterate is in the interior of the nuclear norm ball, the in-face steps often do not lead to rank-drop steps. Moreover, at each iteration when the iterate is in the interior,  a binary search is required to determine the maximum feasible step length once the direction is computed. This requires several SVD updates, leading to very expensive intermediate iterates. Thus, it is critical for the iterates of the in-face method to reach the boundary of the nuclear norm ball quickly. Secondly, the parametrization suggested in \cite{freund2015extended} (i.e. $\gamma_{1} = 0$ and $\gamma_{2} = \infty$) for large datasets corresponds to only taking the maximum step length. On the boundary of the nuclear norm ball, this is equivalent to only accepting iterates with rank decreased. Since our proposed rank-drop is determined optimally to decrease the objective (up to the first order) among all rank-drop steps explicitly, we believe that this is a better optimization formulation when the iterate approaches the boundary.

\subsubsection{Solving the Optimization Problem in the Interior Case}
Now we discuss how to solve the Rank-Drop optimization problem \eqref{prob:rank_drop} in the interior case. 
Let $W \coloneqq U^{\top}\nabla f(X_{k})V$. 

The Lagrangian for \eqref{prob:rank_drop} is,
\begin{align*}
\mathcal{L}(s,t,\lambda, \alpha, \beta) \coloneqq& s^{\top}Wt + \lambda(s^{\top}\Sigma^{-1}t - \kappa(X_{k})^{-1})  \\
&+ \alpha(s^{\top}s - 1) + \beta(t^{\top}t - 1).
\end{align*}

\begin{theorem}
\label{thm:kkt_case1}
Suppose $(s, t, \lambda, \alpha, \beta)$ satisfies the KKT conditions of \eqref{prob:rank_drop} with $\norm{t}_{2} < 1$. Then $\alpha = \beta = 0$, $\lambda$ is an eigenvalue of $-\Sigma W$,  and  $M_{\lambda} \coloneqq -\frac{1}{2}(W + \lambda\Sigma^{-1})$ has zero as a singular value. Conversely, assume that $(\hat{s}, \hat{t})$ forms a singular vector pair of $M_{\lambda}$ associated with the singular value zero and $\kappa(X_{k})\hat{s}^{\top}\Sigma^{-1}\hat{t} > 1$, then   the KKT conditions of \eqref{prob:rank_drop} are satisfied at $(\hat{s},\hat{t}/(\kappa(X_{k})\hat{s}^{\top}\Sigma^{-1}\hat{t}), \lambda, 0, 0)$.
\end{theorem}

Since problem \eqref{prob:rank_drop} is nonconvex and generally difficult to solve, we only consider candidate KKT points characterized in Theorem \ref{thm:kkt_case1}. For each eigenvalue $\lambda$ of  $-\Sigma Wx$, we fix $\alpha = \beta = 0$ and choose $(\hat{s}, \hat{t})$ that corresponds to the singular vector pair of $-0.5(W + \lambda \Sigma^{-1})$ with the singular value zero. If $(s,t) =  (\hat{s},\hat{t}/(\kappa(X_{k})\hat{s}^{\top}\Sigma^{-1}\hat{t}))$ is feasible with respect to \eqref{prob:rank_drop} and $\norm{t}_{2} < 1$, then by Theorem \ref{thm:kkt_case1}, $(s, t, \lambda, 0, 0)$ satisfies the KKT conditions.

In the event that no feasible candidate is found, we  solve \eqref{ann_rank_drop} instead and obtain a feasible rank-drop step, guaranteed by Theorem \ref{thm:ext_feasible}. We remark that, although it is always possible to find feasible rank-drop steps using  \eqref{ann_rank_drop}, in the interior case, it is still preferable to solve problem \eqref{prob:rank_drop} since the assumption that $s = t$ can be restrictive and unnecessary when far away from the boundary. Algorithm \ref{alg:rd_calc} presents the details of the rank-drop direction computation.

\subsection{Step Selection Criteria}
 The criteria in \cite{freund2015extended} for away-steps require solving the regular FW linear subproblem  at each iteration even when the FW step is not taken. In the rank-drop framework, we accept any rank-drop step that does not increase the objective. This allows the algorithm to maintain a lower rank SVD  as well as allows the algorithm to skip computing a FW step unnecessarily. Additionally, for the rank-drop step, we observe that it is very rare that the algorithm chooses a rank-drop step two iterations in a row. Thus, we only generate the rank-drop direction when the previous iterate is a regular FW step, avoiding unnecessary rank-drop step computations.

\section{Convergence Analysis}
Following the proof for Theorem 4 in \cite{guelat1986some}, we  show that the iterates, from  Rank Drop FW in Algorithm \ref{alg:rd_step},  converge to the global optimum of \eqref{prob:nn_reg}.

\begin{theorem}
\label{thm:convergence}
Let $\{X_{k}\}$ be a sequence generated by Algorithm \ref{alg:rd_step} and let $f^{*}$ be the optimal value for problem \eqref{prob:nn_reg}. Assume $\nabla f(X)$ is Lipschitz continuous in the feasible region. Then $ f(X_{k}) - f^{*} \leq \frac{8\delta^{2}L}{4 + N_{fw}^{k}}$, where $N_{fw}^{k}$ be the number of FW steps taken up to the iteration $k$.
\end{theorem}

\begin{algorithm}
  \caption{Compute Rank-Drop Direction (\texttt{rankDrop})}  
  \label{alg:rd_calc}
  \begin{algorithmic}
  \STATE Input: thin SVD $X_{k} \coloneqq U\Sigma V^{\top}$ and $\nabla f(X_{k})$.
  \STATE $\kappa(X_{k}) \gets \frac{\delta - \norm{X_{k}}_{NN}}{2}$
  \STATE $W \gets U^{\top}\nabla f(X_{k})V$
  \IF{$\kappa(X_{k}) \geq \sigma_{r}(X_{k})$}
  \STATE (Interior Case)
  \STATE $\Lambda \gets \texttt{eigs}(-\Sigma W)$
  \STATE $b \gets -\infty $
  \FOR{$\lambda_{i} \in \Lambda$}  
  \STATE $M_{\lambda} \coloneqq -\frac{1}{2}(W + \lambda_{i}\Sigma^{-1})$
  \STATE $(s,t, \sigma) \gets \texttt{SVD}(M_{\lambda})$ (return the zero SV)
  \IF{$\kappa(X_{k})s^{\top}\Sigma^{-1}t \geq 1$ and $ q\left(s, \frac{t}{\kappa(X_{k})s^{\top}\Sigma^{-1}t}\right) > b$}
  \STATE $(s^{*}, t^{*}) \gets (s,t)$
  \STATE $b \gets q\left(s, \frac{t}{\kappa(X_{k})s^{\top}\Sigma^{-1}t}\right)$     
  \ENDIF
  \ENDFOR  
  \STATE $\alpha \gets (s^{\top}\Sigma^{-1}t)^{-1}$
  \STATE $\tau^{*} \gets \frac{\alpha}{\delta - \alpha}$
  \ENDIF
  \IF{$\kappa(X_{k}) < \sigma_{r}(X_{k})$ or $b = -\infty$}
  \STATE (Exterior case or no candidates from the Interior Case)
  \STATE $s^{*} \gets \texttt{genEig}(0.5(W + W^{\top}), \Sigma^{-1})$ 
  \STATE $t^{*} \gets s^{*}$
  \STATE $\tau^{*} \gets (\delta s^{\top}\Sigma^{-1}s - 1)^{-1}$
  \ENDIF 
  \RETURN{$(s^{*}, t^{*}, \tau^{*})$}
    \end{algorithmic}
\end{algorithm}
\begin{algorithm}
  \caption{Rank-Drop Frank-Wolfe (\textbf{RDFW})}  
  \label{alg:rd_step}
  \begin{algorithmic}
  \STATE Let $X_{0} \in \mathcal{S}$, with initial SVD $X_{0} = U\Sigma V^{\top}$, and maximum iteration count $T$
  \STATE $\texttt{prevStep} \gets$ ``RD'' (Start with Frank-Wolfe step)
  \FOR{$k = 0...$}
  \STATE Compute $\nabla f(X_{k})$
  \IF{ \texttt{prevStep} == ``RD''}
  \STATE \texttt{GO TO} (Frank-Wolfe)
  \ENDIF
  \STATE Compute Rank-Drop direction (see Algorithm \ref{alg:rd_calc}): $(s_{k}, t_{k}, \tau^{*}) \gets \texttt{rankDrop}(U, \Sigma, V, \nabla f(X_{k}))$
  \STATE $\tilde{X} \coloneqq X_{k} + \tau^{*} (X_{k} - \delta Us_{k}t_{k}^{\top}V^{\top})$
  \IF{$f(\tilde{X})\leq f(X_{k})$}
  \STATE $X_{k+1} \gets \tilde{X}$
  \STATE $Z_{k} \gets -\delta Us_{k}t_{k}^{\top}V^{\top}$
  \STATE $\texttt{prevStep} \gets$ ``RD''
  \ELSE
  \STATE (Frank-Wolfe)
  \STATE $Z_{k} \gets \argmin_{Z \in \mathcal{B}_{NN}(0, \delta)} \langle Z, \nabla f(X_{k}) \rangle$
  \STATE $\tau^{*} \gets \argmin_{\tau \in [0, 1]} f(X_{k} + \tau (Z_{k} - X_{k}))$
  \STATE $X_{k+1} := X_{k} + \tau^{*}(Z_{k} - X_{k})$ 
  \STATE $\texttt{prevStep} \gets$ ``Frank-Wolfe''
  \ENDIF
  \STATE $(U, \Sigma, V) \gets \texttt{updateSVD}(U, \Sigma, V,  \tau^{*}Z_{k})$ 
  \ENDFOR
\end{algorithmic}
\end{algorithm}

\begin{table*}[!t]
\centering
\footnotesize
\begin{tabular}{llrrrrr}
\specialrule{.2em}{.1em}{.1em}
Dataset & & Frank-Wolfe & Away-Step FW & In-Face$(0, \infty)$ & ActiveALT & Rank-Drop FW\\
\specialrule{.2em}{.1em}{.1em}
  \multirow{3}{*}{ML-100k} & RMSE & 0.878 & 0.878 & 0.878 & 0.877 & 0.879\\
 & Rank (Max) & 501.4 (504) & 466.8 (477) & 54.2 (64) & 43.4 (50) & 41.6 (44) \\
 & Time (s) & 146.09 & 152.49 & 37.85 & 30.28 & 35.45\\
\midrule
\multirow{3}{*}{ML-1M} & RMSE & 0.820 & 0.820 & 0.822 & 0.818 & 0.820\\
 & Rank (Max) & 467.4 (478) & 425.2 (430) & 77.6 (112)& 65 (68)& 67 (69)\\
 & Time (s) & 1,127.70 & 1,197.67 & 356.42 & 503.39 & 284.85\\
\midrule
  \multirow{3}{*}{ML-10M} & RMSE & 0.806 & 0.806 & 0.807 & 0.806 & 0.808 \\
 & Rank (Max) & 1000 (1000) & 889.8 (893) & 210 (297) & 141.8 (144) & 140 (140) \\
 & Time (s) & 41,467.60 & 44,327.57 & 29,630.43 & 13,281.95 & 12,499.43\\
 \midrule
\multirow{3}{*}{ML-20M} & RMSE & - & - & 0.800 & 0.800 & 0.801\\
 & Rank (Max) & - & - & 274.6 (471) & 206.2 (214) & 202 (203) \\ 
 & Time (s) & - & - & 117,535.82 & 38,497.16 & 29,102.62\\
\specialrule{.2em}{.1em}{.1em}
\end{tabular}
\caption{ \small Computational results on matrix completion problems averaged over 5 random initializations. The max rank is the maximum rank observed over all 5 trials. For ML-20M, the FW and AFW algorithms took too long to successfully terminate.}
\end{table*}

\section{Complexity Per Iteration}
When computing the rank-drop steps, we note that the dimension of the subproblems is $r$, the rank of the current iterate. First, forming the matrix $W \coloneqq U^{\top} \nabla f(X_{k})V$ requires $O(r^{2}h\min\{m,n\})$ operations, where $h$ is the maximum number of nonzero elements in any row or column. In the interior case, we must compute an eigen-decomposition of an $r \times r$ matrix which takes $O(r^{3})$ time. Then, each eigenvalue $\lambda$ is used to form the matrix $-0.5(W + \lambda \Sigma^{-1})$ where the singular vector pair corresponding to the zero singular value is computed. The total time required for the interior case is $O(r^{3} + r^{2}h\min\{m,n\})$. In the exterior case, $O(r^{2})$ flops are required to compute the largest eigenvalue. Thus, the total complexity per iteration is $O(r^{3} + r^{2}h\min\{m,n\})$. 

\section{Experimental Results}
\label{experiments}
We  validate the rank-drop steps on a matrix completion task using various datasets from MovieLens\footnote{\url{http://grouplens.org/datasets/movielens/}}.  We first center and scale each data set to have mean 0 and standard deviation 1.
\begin{table}[ht]
\centering
\footnotesize
\begin{tabular}{lccc}
\hline
\textbf{Dataset} & \textbf{\# Users} & \textbf{\# Movies} & \textbf{\# Ratings}\\
\hline
MovieLens 100k & 943 & 1,682 & 100,000\\
MovieLens 1M & 6,040 & 3,900 & 1,000,209\\
MovieLens 10M & 82,248 & 10,681 & 10,000,054\\
MovieLens 20M & 138,493 & 27,278 & 20,000,263\\
\hline
\end{tabular}
\caption{\small MovieLens Data}
\label{tbl:movielens}
\end{table}
We  compare the proposed Rank-Drop Frank Wolfe (RDFW) against the aforementioned FW variants, FW \cite{frank1956algorithm}, AFW \cite{lacoste2015global}, and IF$(0, \infty)$, \cite{freund2015extended}, as well as a state-of-the-art nuclear norm regularized solver in ActiveALT \cite{hsieh2014nuclear}.

Following \cite{yao2016efficient}, we randomly partition each dataset into 50\% training, 25\% validation, and 25\%  testing. The $\delta$ value in \eqref{prob:nn_reg} is tuned with $\delta = \mu_{j} \cdot \norm{Y}_{F}$, where $\norm{Y}_{F}$ is the Frobenius norm of the training data matrix, and $\mu_{j} = 2 + 0.2j, j \in \mathbb{N}$. We increase $j$ until the mean RMSE on the validation set does not improve by more than $10^{-3}$. We terminate the algorithm when an upper bound on the relative optimality gap ensures $(f(X_{k}) - f^{*})/f^{*} < 10^{-2}$ or a maximum iteration count of 1000 is reached.

For ActiveALT, a regularized nuclear norm problem (not constrained) is solved where the regularization parameter $\lambda$ is chosen by approximately solving for the Lagrange multiplier from the solution to the constrained problem. From the optimality conditions, we have $U^{\top}\nabla f(X^{*})V + \lambda I = 0$. Thus, $\lambda$ is approximated by the mean of the diagonal values of $U^{\top} f(X_{k})V$, where $X_{k}$ is the converged solution of RDFW.  ActiveALT terminates when $f(X_{k-1}) - f(X_{k}) < 10^{-4}$, to match with the criterion suggested in \cite{yao2016efficient} with a maximum iteration count of 150. 
\begin{table}[ht]
\footnotesize
\centering
\begin{tabular}{lcc}
\hline
Dataset & Frank-Wolfe $\delta$ & ActiveALT $\lambda$\\
\hline
MovieLens 100k & 3 & 10.94\\
MovieLens 1M & 3.4 & 22.7\\
MovieLens 10M & 5.2 & 49.4\\
MovieLens 20M & 6.6 & 59.04\\
\hline
\end{tabular}
\caption{Parameters used for each dataset.}\label{tb:para}
\end{table}
\subsection{Computational Details}
All simulations were run in MATLAB. For all FW variants, we maintain a thin SVD for the current iterate, where the SVD is updated at each iteration using a rank-one update as described in \cite{brand2006fast}. The rank is calculated by counting all singular values larger than $10^{-6}$, in agreement with the definition of rank in \cite{freund2015extended}.

\section{Discussion} 
We observe that the proposed method, RDFW,  shows large improvements over  FW and its common variants in terms of the rank and computing time. While ActiveALT, the leading nuclear norm based matrix completion solver, is very competitive with RDFW,  RDFW has the additional attractive property that it does not require knowledge of the structure of the Hessian, or even require $f(\cdot)$ to be twice differentiable, allowing for greater generality. Moreover, since the matrix completion objective is quadratic, this is the ideal situation for ActiveALT since the underlying solver in ActiveALT only requires one Newton step to converge, whereas with a nonquadratic objective, the computational burden can increase. 

\section{Conclusions}
We have proposed a rank-drop optimization formulation to determine optimally descent rank-drop steps for the nuclear-norm constrained minimization. By considering the interior and exterior cases separately to ensure feasibility, we also devise subproblems that can be efficiently solved.
The proposed formulation can be deployed in a projection free minimization method, e.g., FW method, to efficiently compute a low rank solution by maintaining low rank intermediate iterates, without compromising the strong convergence guarantees. While classic FW methods tend to have very high rank solutions for nuclear-norm constrained problems, we have shown that the addition of rank-drop steps can drastically reduce the rank of the iterates, allowing for much faster algorithms to reach low rank solutions with less required space.

\section*{Acknowledgements}
Authors acknowledge funding from the National Sciences and Engineering Research Council of Canada.

\newpage
\bibliographystyle{named}
\bibliography{main}

\begin{thebibliography}{}

\bibitem[\protect\citeauthoryear{Brand}{2006}]{brand2006fast}
Matthew Brand.
\newblock Fast low-rank modifications of the thin singular value decomposition.
\newblock {\em Linear algebra and its applications}, 415(1):20--30, 2006.

\bibitem[\protect\citeauthoryear{Chu \bgroup \em et al.\egroup
  }{1998}]{chu1998rank}
Moody~T Chu, RE~Funderlic, and Gene~H Golub.
\newblock Rank modifications of semidefinite matrices associated with a secant
  update formula.
\newblock {\em SIAM Journal on Matrix Analysis and Applications},
  20(2):428--436, 1998.

\bibitem[\protect\citeauthoryear{Egerv{\'a}ry}{1960}]{egervary1960rank}
Eugen Egerv{\'a}ry.
\newblock On rank-diminishing operations and their applications to the solution
  of linear equations.
\newblock {\em Zeitschrift f{\"u}r angewandte Mathematik und Physik ZAMP},
  11(5):376--386, 1960.

\bibitem[\protect\citeauthoryear{Fazel \bgroup \em et al.\egroup
  }{2001}]{fazel2001rank}
Maryam Fazel, Haitham Hindi, and Stephen~P Boyd.
\newblock A rank minimization heuristic with application to minimum order
  system approximation.
\newblock In {\em American Control Conference, 2001. Proceedings of the 2001},
  volume~6, pages 4734--4739. IEEE, 2001.

\bibitem[\protect\citeauthoryear{Frank and Wolfe}{1956}]{frank1956algorithm}
Marguerite Frank and Philip Wolfe.
\newblock An algorithm for quadratic programming.
\newblock {\em Naval research logistics quarterly}, 3(1-2):95--110, 1956.

\bibitem[\protect\citeauthoryear{Freund \bgroup \em et al.\egroup
  }{2015}]{freund2015extended}
Robert~M Freund, Paul Grigas, and Rahul Mazumder.
\newblock An extended frank-wolfe method with" in-face" directions, and its
  application to low-rank matrix completion.
\newblock {\em arXiv preprint arXiv:1511.02204}, 2015.

\bibitem[\protect\citeauthoryear{Gu{\'e}lat and
  Marcotte}{1986}]{guelat1986some}
Jacques Gu{\'e}lat and Patrice Marcotte.
\newblock Some comments on wolfe's ‘away step’.
\newblock {\em Mathematical Programming}, 35(1):110--119, 1986.

\bibitem[\protect\citeauthoryear{Hsieh and Olsen}{2014}]{hsieh2014nuclear}
Cho-Jui Hsieh and Peder~A Olsen.
\newblock Nuclear norm minimization via active subspace selection.
\newblock In {\em ICML}, pages 575--583, 2014.

\bibitem[\protect\citeauthoryear{Jaggi}{2013}]{jaggi2013revisiting}
Martin Jaggi.
\newblock Revisiting frank-wolfe: Projection-free sparse convex optimization.
\newblock In {\em ICML (1)}, pages 427--435, 2013.

\bibitem[\protect\citeauthoryear{Lacoste-Julien and
  Jaggi}{2015}]{lacoste2015global}
Simon Lacoste-Julien and Martin Jaggi.
\newblock On the global linear convergence of frank-wolfe optimization
  variants.
\newblock In {\em Advances in Neural Information Processing Systems}, pages
  496--504, 2015.

\bibitem[\protect\citeauthoryear{Watson}{1992}]{watson1992characterization}
G~Alistair Watson.
\newblock Characterization of the subdifferential of some matrix norms.
\newblock {\em Linear algebra and its applications}, 170:33--45, 1992.

\bibitem[\protect\citeauthoryear{Wolfe}{1970}]{wolfe1970convergence}
Philip Wolfe.
\newblock Convergence theory in nonlinear programming.
\newblock {\em Integer and nonlinear programming}, pages 1--36, 1970.

\bibitem[\protect\citeauthoryear{Yao \bgroup \em et al.\egroup
  }{2016}]{yao2016efficient}
Quanming Yao, James Kwok, et~al.
\newblock Efficient learning with a family of nonconvex regularizers by
  redistributing nonconvexity.
\newblock {\em arXiv preprint arXiv:1606.03841}, 2016.

\end{thebibliography}

\iffull
\appendix
\section{Omitted Proofs}
\subsection{Proof of Lemma \ref{lem:rewrite_rd}}
\begin{proof}
Assume that $\sigma^{-1} u v^T \in \mathcal{R}(A)$. Then
 $u = Ax$ and $v = A^{\top}y$ for some $x$ and $y$ and $u,v \neq 0$.  Let $A = U\Sigma V^{\top}$ be a thin SVD and write $u = Us$ and  $v = Vt$ for some $s, t \in \mathbb{R}^{r}$. 

We have,
\begin{align}
Vt = A^{\top}y = V\Sigma U^{\top}y \Rightarrow \Sigma^{-1}t &= U^{\top}y.
\end{align}
Consequently,
\begin{equation}
\label{id:replace_bil}
\sigma = y^{\top}Ax = y^{\top}(Us) = (U^{\top}y)^{\top}s = t^{\top}\Sigma^{-1}s.
\end{equation}

Finally, we note that $t^{\top}\Sigma^{-1}s$ can always be made positive by replacing $s$ with $-s$.
\end{proof}

\subsection{Proof of Lemma \ref{lem:radius_bound}}
\begin{proof}
Let $Z = \sigma^{-1}uv^{\top}$. Without loss of generality, we can assume that $\sigma^{-1} > 0$ since we can always substitute $u$ with $-u$ to make $\sigma^{-1}$ positive. Let $\alpha = \sigma^{-1}\norm{u}_2\norm{v}_2$. We need to show that $\alpha \geq \sigma_{r}(A)$.

From Lemma \ref{lem:rewrite_rd}, we know that $\sigma^{-1}uv^{\top} = \frac{Ust^{\top}V^{\top}}{s^{\top}\Sigma^{-1}t}$ for some $s, t \in \mathbb{R}^{r}$. Then,
\begin{align*}
\begin{split}
\sigma^{-1}\norm{u}_{2}\norm{v}_{2} = \frac{\norm{Us}_{2}\norm{Vt}_{2}}{\abs{s^{\top}\Sigma^{-1}t}} &\geq \frac{\norm{s}_{2}\norm{t}_{2}}{\norm{s}_{2}\norm{\Sigma^{-1}t}_{2}}\\
&\geq \frac{1}{\max_{z: \norm{z}_{2} = 1}\norm{\Sigma^{-1}z}_{2}}\\
&= \sigma_{r}(A)
\end{split}
\end{align*}
\end{proof}

\subsection{Proof of Theorem \ref{thm:maintain_cvx}}
\begin{proof}
Define $R_{k} = X_{k} - Z_{k}$. Note that,
\begin{align*}
\norm{R_{k}}_{NN} + \norm{Z_{k}}_{NN} &= \norm{X_{k} - Z_{k}}_{NN} + \norm{Z_{k}}_{NN}\\
&\leq \norm{X}_{k} + 2\norm{Z_{k}}_{NN}\\
& \leq \delta
\end{align*}
where the last inequality uses $\norm{Z_{k}}_{NN} \leq (\delta - \norm{X_{k}}_{NN})/2$.

Let $R_{k}$ have the thin SVD, $R_{k} = \sum_{i=1}^{r-1}\sigma_{i}u_{i}v_{i}^{\top}$. We can express $X_{k}$ as,
\begin{equation}
\label{eq:cvx_combo}
X_{k} = \alpha_{0}\delta \hat{Z}_{k} + \sum_{i=1}^{r-1}\alpha_{i}\delta u_{i}v_{i}^{\top}
\end{equation}
where $\alpha_{0} = \norm{Z_{k}}_{NN}/\delta$ and $\alpha_{i} = \sigma_{i}/\delta$ for $i \in \{1,...,r-1\}$.
Note that $\alpha_{i} \geq 0,~i=0,\ldots, r-1$, and
\begin{align}
\begin{split}
    \norm{R_{k}}_{NN} + \norm{Z_{k}}_{NN} &\leq \delta\\
    \Rightarrow \delta\left(\sum_{i=1}^{r-1}\alpha_{i}\right) + \delta\alpha_{0} &\leq \delta\\
    \Rightarrow \sum_{i=0}^{r-1}\alpha_{i} &\leq 1
\end{split}
\end{align}

Next, we wish to show that $X_{k+1} \in \mathcal{B}_{NN}(0, \delta)$. Rewrite $X_{k+1}$ as,
\begin{align*}
X_{k+1} &= X_{k} + \tau X_{k} - \tau\delta\hat{Z}_{k}\\
&= (1 + \tau)R_{k} + (1 + \tau)\alpha_{0}\delta \hat{Z}_{k} - \tau\delta\hat{Z}_{k} ~~ \text{(using \eqref{eq:cvx_combo})}\\
&= (1 + \tau)R_{k} + (\alpha_{0} + \alpha_{0}\tau - \tau)\delta\hat{Z}_{k}\\
&= \sum_{i=1}^{r-1}(1+\tau)\alpha_{i}\delta u_{i}v_{i}^{\top} + (\alpha_{0} + \alpha_{0}\tau - \tau)\delta \hat{Z}_{k}\\
&= \sum_{i=1}^{r-1}\beta_{i}\delta u_{i}v_{i}^{\top} + \beta_{0}\delta \hat{Z}_{k}
\end{align*}
where $\beta_{0} = (\alpha_{0} + \alpha_{0}\tau - \tau)$ and $\beta_{i} = (1+\tau)\alpha_{i}, \forall i \in \{1,...,r-1\}$. 

If $\tau \in [0, \tau^{*}]$, then $\alpha_{0} + \alpha_{0}\tau - \tau \geq 0$, with equality at $\tau = \tau^{*}$. Thus, $\beta_{i} \geq 0, \forall i \in \{0,1,...,r-1\}$. Next, we have that,
\begin{align*}
    \sum_{i=0}^{r-1}\beta_{i} &= (1+ \tau)\sum_{i=1}^{r-1}\alpha_{i} + (1+\tau)\alpha_{0} - \tau\\
    &= (1+ \tau)\sum_{i=0}^{r-1}\alpha_{i} - \tau\\
    &\leq (1 + \tau) - \tau\\
    &= 1.
\end{align*}
Thus, $\beta_{i} \geq 0,~i=0,\ldots,r-1$, and $\sum_{i=0}^{r-1}\beta_{i} \leq 1$. This implies that $X_{k+1} \in \mathcal{B}(0, \delta)$ if $\tau \in [0, \tau^{*}]$.

Finally, when $\tau = \tau^{*}$, we have that $(\alpha_{0} + \alpha_{0}\tau^{*} - \tau^{*}) = 0$. Hence, $X_{k+1}$ is a sum of $r-1$ rank one matrices and $rank(X_{k+1}) \leq r-1$. Since $X_{k+1}$ is a rank one perturbation of $X_{k}$, the most the rank can decrease by is 1, thus  $rank(X_{k+1}) = r-1$.
\end{proof}

\subsection{Proof of Theorem \ref{thm:feasible}}
\begin{proof}
Assume that $X_k$ has the thin SVD $X_k = U\Sigma V^T$.
We show that the smallest singular vector pair can be made into a rank-drop step. Let $s = e_{r}$, where $e_{r}$ is the $r$\textsuperscript{th} elementary basis vector,  $t = (\sigma_{r}(X_{k})/\kappa(X_{k}))e_{r}$, and $Z = Ust^{\top}V^{\top}/s^{\top}\Sigma^{-1}t$. Since $\sigma_{r}(X_{k}) \leq \kappa(X_{k})$,  $\norm{t}_{2} \leq 1$. We  conclude that,
\begin{align}
    \begin{split}
    \norm{Z}_{NN} &= \norm*{\frac{Ust^{\top}V^{\top}}{s^{\top}\Sigma^{-1}t}}_{NN}\\
    &\leq \norm{Us}_{2}\norm{Vt}_{2}(\norm{\Sigma^{-1}s}_{2}\norm{t}_{2})^{-1}\\
    &\leq \sigma_{r}(X_{k})\\
    &\leq \kappa(X_{k})
    \end{split}
\end{align}
Thus $Z \in \mathcal{B}_{NN}(0, \kappa(X_{k})) \cap \mathcal{R}(X_{k})$, and the feasible region to \eqref{intCase} is non-empty. More generally, we can see that any singular vector pair with singular value $\sigma_{i}(X_{k}) \leq \kappa(X_{k})$ will be feasible to \eqref{intCase} as well.
\end{proof}

\subsection{Proof of Theorem \ref{thm:equv}}

\begin{proof}
First we will show that for any feasible solution $(s,t)$ to \eqref{prob:fractional}, there exists a corresponding feasible solution to \eqref{prob:rank_drop} with the same objective value.

Let $(s,t)$ be a feasible solution to \eqref{prob:fractional}. Define $\hat{s} \coloneqq \frac{s}{\norm{s}_{2}} \text{ and } \hat{t} \coloneqq \frac{\kappa(X_{k})^{-1}}{\hat{s}^{\top}\Sigma^{-1}t} t$. We will show $(\hat{s}, \hat{t})$ is feasible for \eqref{prob:rank_drop} and the objective values of \eqref{prob:fractional} and \eqref{prob:rank_drop} are equal at $(s,t)$ and $(\hat{s}, \hat{t})$. To see this, note that,
\begin{align}
    \begin{split}
        \frac{Ust^{\top}V^{\top}}{s^{\top}\Sigma^{-1}t} &= \frac{\frac{\kappa(X_{k})^{-1}}{\norm{s}_{2}\hat{s}^{\top}\Sigma^{-1}_{k}t}}{\frac{\kappa(X_{k})^{-1}}{\norm{s}_{2}\hat{s}^{\top}\Sigma^{-1}_{k}t}}\cdot \frac{Ust^{\top}V^{\top}}{s^{\top}\Sigma^{-1}t}\\
        &= \frac{U\hat{s}\hat{t}^{\top}V^{\top}}{\hat{s}^{\top}\Sigma^{-1}\hat{t}}\\
    \end{split}
\end{align}
But,
\begin{equation}
\hat{s}^{\top}\Sigma^{-1}\hat{t} = \frac{\kappa(X_{k})^{-1}}{\hat{s}^{\top}\Sigma^{-1}t}\hat{s}^{\top}\Sigma^{-1}t = \kappa(X_{k})^{-1},
\end{equation}
satisfying the first constraint of \eqref{prob:rank_drop}. We can also conclude,
\begin{equation}
\label{id:equiv_rep}
\frac{Ust^{\top}V^{\top}}{s^{\top}\Sigma^{-1}t} = \kappa(X_{k}) U\hat{s}\hat{t}^{\top}V^{\top}
\end{equation}
showing the objective values will be equal.

For the norm constraint, $\norm{\hat{s}}_{2} = 1$ by construction. Next, to see that the solution satisfies the last constraint, the fact that $(s,t)$ is feasible gives us,
\begin{align}
\begin{split}
\label{id:subst}
\norm*{\frac{Ust^{\top}V^{\top}}{s^{\top}\Sigma^{-1}t}}_{NN} &\leq \kappa(X_{k}) \\
\Rightarrow \kappa(X_{k})\norm{U\hat{s}\hat{t}^{\top}V^{\top}}_{NN} &\leq \kappa(X_{k})\\
\Rightarrow \norm{U\hat{s}\hat{t}^{\top}V^{\top}}_{NN} &\leq 1
\end{split}
\end{align}
where the first implication uses the result from \eqref{id:equiv_rep}. Next, using the fact that the nuclear norm of a rank-one matrix is equivalent to its Frobenius norm, we have,

\begin{align}
    \label{id:norm_reduce}
    \begin{split}
        \norm{U\hat{s}\hat{t}^{\top}V^{\top}}_{NN} &= \norm{U\hat{s}\hat{t}^{\top}V^{\top}}_{F}\\
        &= \sqrt{\text{tr}((U\hat{s}\hat{t}^{\top}V^{\top})^{\top}(U\hat{s}\hat{t}^{\top}V^{\top}))}\\
        &= \sqrt{\text{tr}(V\hat{t}\hat{s}^{\top}U^{\top}U\hat{s}\hat{t}V^{\top})}\\
        &= \sqrt{\text{tr}(V\hat{t}\hat{t}^{\top}V^{\top}})\\
        &= \sqrt{\text{tr}(\hat{t}^{\top}V^{\top}V\hat{t}})\\
        &= \norm{\hat{t}}_{2}
    \end{split}
\end{align}
Thus, the results from \eqref{id:subst} and \eqref{id:norm_reduce} jointly imply that $\norm{\hat{t}}_{2} \leq1$. It is readily seen that any feasible solution to \eqref{prob:rank_drop} is a feasible solution to \eqref{prob:fractional}. Since there exists a mapping from feasible points in \eqref{prob:fractional} to \eqref{prob:rank_drop} and vice versa preserving objective values, the optimal values must be equal. Thus, an optimal solution to \eqref{prob:rank_drop} is an optimal solution to \eqref{prob:fractional} and the converse result holds as stated.
\end{proof}

\subsection{Proof of Theorem \ref{thm:descent}}
\begin{proof}
From \cite{watson1992characterization}, the subdifferential of the nuclear norm is,
\begin{equation*}
    \partial \norm{X_{k}}_{NN} \coloneqq \{ UV^{\top} + H : U^{\top}H = HV = 0, \norm{H}_{sp} \leq 1\}.
\end{equation*}
Let $G \in \partial\norm{X_{k}}_{NN}$ be an arbitrary subgradient. Then,
\begin{align*}
&\phantom{{}={}}\langle G, X_{k} - \delta Ust^{\top}V^{\top} \rangle\\
&= \langle UV^{\top} + H, U\Sigma V^{\top} - \delta Ust^{\top}V^{\top} \rangle\\
&=  \text{tr}((UV^{\top}+H)^{\top}(U\Sigma V^{\top} - \delta Ust^{\top}V^{\top})\\
&= \text{tr}( VU^{\top}U\Sigma V^{\top} + H^{\top}U\Sigma V^{\top}) -\\ &\phantom{{}={}}\delta\text{tr}(VU^{\top}Ust^{\top}V^{\top} + H^{\top}Ust^{\top}V^{\top})\\
&= \text{tr}(\Sigma) - \delta s^{\top}t\\
&= \norm{X_{k}}_{NN} - \delta s^{\top}t
\end{align*}
This implies that $\max_{G \in \partial \norm{X}_{NN}} \langle D_{k}, G \rangle \leq 0$  if and only if $ \delta s^{\top}t \geq \norm{X_{k}}_{NN}$. Since $D_{k}$ has a negative inner product with all elements in the subdifferential, it must be a descent direction for the nuclear norm at $X_{k}$.
\end{proof}

\subsection{Proof of  Corollary \ref{cor:sym}}
\begin{proof}
From Theorem \ref{thm:descent}, we must have that $\delta s^{\top}t \geq \norm{X_{k}}_{NN} = \delta$. This implies that $s^{\top}t \geq 1$. Since $\norm{s}_{2} = \norm{t}_{2} = 1$, we have that $s^{\top}t \leq 1$, where equality is attained only when $s = t$. Thus, $\delta s^{\top}t \geq \norm{X_{k}}_{NN}$ if and only if $s = t$, completing the proof.
\end{proof}

\subsection{Proof of Lemma \ref{lm:bdnn}}

\begin{proof}
Note that $\sigma_{r}(X_{k}) < \norm{X_{k}}_{NN}/r$. Then, the inequality can be rearranged as follows.
\begin{align*}
\kappa(X_{k}) & < \sigma_{r}(X_{k})\\
\delta - \norm{X_{k}}_{NN} &< 2\frac{\norm{X_{k}}_{NN}}{r}\\
\norm{X_{k}}_{NN} & > \frac{r}{r+2}\delta
\end{align*}
\end{proof}

\subsection{Proof of Theorem \ref{thm:ext_feasible}}
\begin{proof}
From the definition of $X_{k+1} = (1+\tau^{*})U\Sigma V^{\top} - \tau^{*}\delta Uss^{\top}V^{\top} = U( (1+\tau^{*})\Sigma - \tau^{*}\delta ss^{\top}) V^{\top}$. Let $M = (1+\tau^{*})\Sigma - \tau^{*}\delta ss^{\top}$. It is clear that $rank(X_{k+1}) = rank(M)$. From Theorem \ref{thm:rank_drop}, it is straightforward to verify that $\tau^{*}\delta ss^{\top} \in \mathcal{R}((1+\tau^{*})\Sigma)$, so $rank(M) = rank(X_{k}) - 1$.

Let $\lambda_{i}(M)$ be the $i\textsuperscript{th}$ eigenvalue of $M$, then,
\begin{align}
\begin{split}
\sum_{i}^{r}\lambda_{i}(M) &= \text{tr}(M)\\
&= (1+\tau^{*})\text{tr}(\Sigma) - \tau^{*}\text{tr}(\delta ss^{\top})\\
&\leq \delta. 
\end{split}
\end{align}

We require the following Theorem from \cite{chu1998rank}.

\begin{theorem}[\cite{chu1998rank}]
\label{thm:chu}
Suppose that $D$ is symmetric positive semidefinite, $S$ is symmetric, and \\
$rank(D - S) = rank(D)-rank(S)$. Then $D - S$ is positive semidefinite.
\end{theorem}

From Theorem \ref{thm:chu}, $M$ is symmetric positive semidefinite, so $\sum_{i}\lambda_{i}(M) = \sum_{i}\sigma_{i}(M) \Rightarrow \norm{M}_{NN} \leq \delta$. Since $X_{k+1} = UMV^{\top}$, it follows that $\norm{X_{k+1}}_{NN} \leq \delta$.
\end{proof}

\subsection{Proof of Lemma \ref{lem:int_if}}
\begin{proof}
If $\kappa(X_{k}) > \sigma_{r}(X_{k})$, then $\norm{X_{k}}_{NN} < \delta$, and $\mathcal{F}(X_{k}) = \mathcal{B}_{NN}(0, \delta)$. Then $Z = \kappa(X_{k}) Ust^{\top}V^{\top}$ where $\norm{s}_{2} = 1$ and $\norm{t}_{2} \leq 1$. This implies that $\norm{Z}_{NN} \leq \kappa(X_{k}) < \delta$ and $Z \in \mathcal{F}(X_{k})$.
\end{proof}

\subsection{Proof of Lemma \ref{lem:ext_if}}

\begin{proof}
If $s$ is a solution to \eqref{ann_rank_drop}, then $\norm{s}_{2} = 1$ and $\norm{Z}_{NN} \leq \delta$. If $\norm{X_{k}}_{NN} < \delta$, then $Z \in \mathcal{B}_{NN}(0, \delta) = \mathcal{F}(X_{k})$. 

Otherwise, $\norm{X_{k}}_{NN} = \delta$ and $\mathcal{F}(X_{k}) = \{UMV^{\top}: M \succeq 0, \text{tr}(M) = \delta\}$, as defined in \eqref{def:minimal_face}. Clearly, $\delta ss^{\top}$ is symmetric positive semidefinite, and since $\text{tr}(\delta ss^{\top}) = \delta s^{\top}s = \delta$, we can set $M = \delta ss^{\top}$ and $Z = UMV^{\top} \in \mathcal{F}(X_{k})$. 
\end{proof}

\subsection{Proof of Lemma \ref{thm:kkt_case1}}

\begin{proof}
Since $(s,t,\lambda, \alpha,\beta)$ satisfies the KKT conditions, we have,
\begin{align}
    \begin{split}
        \label{stationarity}
        W^{\top}s + \lambda\Sigma^{-1}s &= -2\beta t\\
        Wt + \lambda\Sigma^{-1}t &= -2\alpha s.\\
    \end{split}
\end{align}

Let $M_{\lambda} = -\frac{1}{2}(W + \lambda\Sigma^{-1})$. Then we can rewrite \eqref{stationarity} as,
\begin{align}
    \begin{split}
        \label{stationarity2}
        M_{\lambda}^{\top}s &= \beta t\\
        M_{\lambda}t &= \alpha s.
    \end{split}
\end{align}
Following these equations, we  conclude that $\alpha = s^{\top}M_{\lambda}t = \beta \norm{t}^{2}_{2}$ for any feasible solution, since $s^{\top}s = 1$. 

Since $(s,t,\lambda, \alpha, \beta)$ satisfies the KKT conditions, from the complementary slackness and the additional assumption $t^{\top}t < 1$, we must have $\beta = 0$. This implies that $0 = \beta = \alpha$, since $\alpha = \beta\norm{t}^{2}$. Thus, $M_{\lambda}$ must be rank deficient.
In other words, there exists $\lambda \in \mathbb{R}$ and a vector $x \in \mathbb{R}^{r}$ such that,
\begin{equation*}
    (W + \lambda\Sigma^{-1})x = 0 \Leftrightarrow -\Sigma Wx = \lambda x.
\end{equation*}
Hence $\lambda$ is an eigenvalue  of $-\Sigma W$. 

With $\alpha = \beta = 0$, for each real $\lambda \in \text{eigs}(-\Sigma W)$, \eqref{stationarity} is satisfied by setting $\hat{s}$ and $\hat{t}$ to any singular value pair of $M_{\lambda}$ with associated singular value of 0. 
Conversely, assume that $(\hat{s}, \hat{t})$ forms a singular vector pair of $-\frac{1}{2}(W + \lambda\Sigma^{-1})$ associated with the singular value zero, and $\kappa(X_{k})\hat{s}^{\top}\Sigma^{-1}\hat{t} > 1$. Then $(s,t) = (\hat{s},\hat{t}/(\kappa(X_{k})\hat{s}^{\top}\Sigma^{-1}\hat{t}))$ is a feasible point of \eqref{prob:rank_drop} and $(s, t, \lambda, 0, 0)$ satisfies the KKT condition of \eqref{prob:rank_drop}.
Thus, we have characterized all possible points that can satisfy the KKT conditions such that $t^{\top}t < 1$.
\end{proof}

\subsection{Proof of Theorem \ref{thm:convergence}}

\begin{proof}
We will use a similar proof as \cite{guelat1986some}. Since rank-drop steps always decrease the rank of the solution, the number of rank-drop steps is bounded by the number of Frank-Wolfe steps. Thus, any sequence $\{X_{k}\}$ contains an infinite number of Frank-Wolfe steps. Since rank-drop steps can only decrease the objective, the convergence is guaranteed by the same arguments as the regular Frank-Wolfe algorithm.
\end{proof}
\fi
\end{document}